\documentclass[letterpaper]{article} 
\usepackage{aaai25}  
\usepackage{times}  
\usepackage{helvet}  
\usepackage{courier}  
\usepackage[hyphens]{url}  
\usepackage{graphicx} 
\usepackage{natbib}  
\usepackage{caption} 
\usepackage{algorithm}
\usepackage{algorithmic}

\usepackage{newfloat}
\usepackage{listings}
\DeclareCaptionStyle{ruled}{labelfont=normalfont,labelsep=colon,strut=off} 
\floatstyle{ruled}
\newfloat{listing}{tb}{lst}{}
\floatname{listing}{Listing}
\title{Fairness of Automatic Speech Recognition: Looking Through a Philosophical Lens}
\author {
    Anna Seo Gyeong Choi\textsuperscript{\rm 1},
    Hoon Choi\textsuperscript{\rm 2}
}
\affiliations {
    \textsuperscript{\rm 1}Department of Information Science, Cornell University, USA\\
    \textsuperscript{\rm 2}Division of Liberal Studies, Kangwon National University, South Korea\\
    sc2359@cornell.edu, choih@kangwon.ac.kr 
}

\usepackage{bibentry}

\begin{document}

\maketitle

\begin{abstract}
Automatic Speech Recognition (ASR) systems now mediate countless human-technology interactions, yet research on their fairness implications remains surprisingly limited. This paper examines ASR bias through a philosophical lens, arguing that systematic misrecognition of certain speech varieties constitutes more than a technical limitation -- it represents a form of disrespect that compounds historical injustices against marginalized linguistic communities. We distinguish between morally neutral classification (\emph{discriminate\textsubscript{1}}) and harmful discrimination (\emph{discriminate\textsubscript{2}}), demonstrating how ASR systems can inadvertently transform the former into the latter when they consistently misrecognize non-standard dialects.
We identify three unique ethical dimensions of speech technologies that differentiate ASR bias from other algorithmic fairness concerns: the temporal burden placed on speakers of non-standard varieties (``temporal taxation''), the disruption of conversational flow when systems misrecognize speech, and the fundamental connection between speech patterns and personal/cultural identity. These factors create asymmetric power relationships that existing technical fairness metrics fail to capture.
The paper analyzes the tension between linguistic standardization and pluralism in ASR development, arguing that current approaches often embed and reinforce problematic language ideologies. We conclude that addressing ASR bias requires more than technical interventions; it demands recognition of diverse speech varieties as legitimate forms of expression worthy of technological accommodation. This philosophical reframing offers new pathways for developing ASR systems that respect linguistic diversity and speaker autonomy.
\end{abstract}

%

\section{Introduction}

In recent years, Automatic Speech Recognition (ASR) has become a mainstay of everyday life. From virtual assistants on smartphones to corporate call centers, voice-based systems now mediate a significant portion of human-technology interactions. Most of these systems rely on machine learning (ML) algorithms that implement inductive reasoning: they analyze patterns in large speech datasets and extrapolate from these patterns to interpret new audio inputs. Although this approach has led to notable improvements in accuracy, it is also vulnerable to biases embedded in the historical data and design choices that shape these models.

A growing body of research has documented biases in ASR, particularly around racial, ethnic, and dialect distinctions, showing that models often perform substantially worse on speakers whose voice characteristics deviate from a presumed ``standard'' \cite{koenecke2020racial, zhao2024quantification, ngueajio2022hey}. \citet{koenecke2020racial} found that commercial ASR systems misrecognize speakers of African American Vernacular English (AAVE) at nearly twice the rate of white speakers, while similar disparities exist for d/Deaf and Hard of Hearing speakers \cite{zhao2024quantification}. 

Crucially, these performance gaps are more than just a reflection of insufficient data or technical glitches. When an ASR system persistently misrecognizes or mistranscribes certain voices, it can restrict access to services, marginalize communities, and implicitly convey the message that some voices matter less than others. Moreover, the temporal nature of speech creates unique ethical implications: unlike text-based systems, ASR errors interrupt conversational flow, impose repetition burdens on certain speakers, and create asymmetric power dynamics that existing fairness frameworks struggle to capture. Over time, such biases risk reinforcing social hierarchies that have historically disadvantaged minority linguistic groups. 

Against this backdrop, scholars and practitioners have proposed a variety of technical interventions, for example, developing more balanced datasets, refining acoustic models, or creating fairness metrics tailored to speech \cite{dheram2022toward, swain2024mitigating, veliche2023improving}. These strategies are essential for mitigating performance disparities and fostering more equitable ASR systems. However, this paper argues that a purely technical focus is insufficient for grasping the deeper moral and political implications of ASR biases. We contend that philosophical analysis can shed critical light on why certain statistical practices become pernicious forms of discrimination. Specifically, we adopt the concept of `disrespect' to pinpoint when inductive inferences, although perhaps efficient or statistically defensible in the aggregate, can still be ethically indefensible when applied to people who speak a different dialect, accent, or language variety.

In foregrounding a philosophical lens, we depart from many fairness studies that concentrate on quantitative disparities in algorithmic performance (e.g., in the context of race or gender \cite{koenecke2020racial, tatman2017gender, martin2020understanding}). While such metrics remain essential, they often do not capture the qualitative harm inflicted by misrecognition. Indeed, scholars working on bias in criminal justice algorithms (e.g., COMPAS) generally acknowledge that racial or gender-based discrimination is ethically wrong and legally prohibited \cite{lagioia2023algorithmic, angwin2022bias} -- yet the underlying philosophical reasons why algorithmic discrimination is wrong can remain unarticulated or under-examined. We extend these insights to ASR, a domain where sociolinguistic-based discrimination is comparatively understudied, yet can be equally insidious and impactful \cite{holliday2021perception}. 

This paper makes several contributions. First, it examines how bias in ASR differs fundamentally from other forms of algorithmic discrimination through its temporal dimensions and direct impact on linguistic identity. Second, it applies philosophical concepts of discriminate\textsubscript{1} (morally neutral sorting) and discriminate\textsubscript{2} (ethically problematic discrimination) to clarify when ASR performance disparities constitute disrespect. Third, it analyzes the tension between linguistic standardization and pluralism in ASR development. By situating biases in ASR within broader questions of moral status, autonomy, and equality, we not only highlight the gravity of these issues but also propose new avenues for designing solutions that genuinely respect linguistic diversity.

In uniting philosophical reflection with empirical case analysis, our hope is to enrich ongoing conversations about fairness in technology. Voice interfaces promise unprecedented convenience and inclusivity; yet if they fail to honor the linguistic and cultural pluralism of their users, these same interfaces risk aggravating social divisions. Grounded in moral and political philosophy, the arguments herein aim to ensure that the accelerating adoption of ASR does not come at the expense of respect, dignity, and equality for all speakers. 

This philosophical reframing matters because it fundamentally changes how we approach solutions. Technical fixes that improve aggregate accuracy metrics may still perpetuate disrespect if they fail to address the temporal burdens and identity erasure experienced by marginalized speakers. By distinguishing between discriminate$_1$ and discriminate$_2$, we can identify when seemingly neutral technical decisions cross ethical boundaries -- knowledge essential for policymakers, designers, and legal frameworks attempting to regulate voice technologies. Without this philosophical foundation, even well-intentioned technical improvements risk reinforcing the very hierarchies they claim to address.

\section{Philosophical Foundations of Fairness and Discrimination}\label{sec2}

\subsection{Inductive Reasoning and Statistical Discrimination}

ML systems, including ASR, fundamentally rely on inductive reasoning -- drawing generalized conclusions from specific instances \cite{harris2021induction}. Unlike deductive reasoning, induction involves inherent uncertainty: models derive patterns from historical data that may not fully represent real-world diversity \cite{copi2016introduction}. This uncertainty becomes ethically significant when systematic imbalances in training data lead to consistent misrecognition of certain speech patterns, particularly those associated with marginalized communities. Even if these errors appear statistically benign when measured by aggregate metrics, they can have profound implications when concentrated among specific demographic groups.

\subsection{From Neutral Classification to Harmful Discrimination}

ML systems necessarily engage in classification (\emph{discriminate\textsubscript{1}}), but this morally neutral sorting can become harmful bias (\emph{discriminate\textsubscript{2}}) under specific conditions. The distinction helps us identify when ASR performance disparities transition from technical limitations to ethical problems. While statistical generalizations are unavoidable in both human reasoning and ML systems, they become problematic when they reinforce social hierarchies, particularly for historically marginalized groups.

Hellman's concept of compounding injustice \cite{hellman2023big} provides a useful framework: harm accumulates when algorithmic decisions build upon existing inequalities, layering new disadvantages over historical ones. For instance, an ASR system that consistently misrecognizes AAVE speakers doesn't merely produce incorrect transcriptions -- it compounds historical marginalization of these speech communities by restricting their access to voice-mediated services and technologies. This perspective aligns with Lippert-Rasmussen's warnings about algorithms operating in ``unjust worlds'' that inherit and magnify preexisting biases \cite{lippert2023using}.

\subsection{Disrespect as the Core Ethical Problem}\label{sec2.3}

The concept of ``disrespect'' offers a precise criterion for distinguishing harmful discrimination from neutral classification in ASR systems. For example, when an ASR system consistently achieves 95\% accuracy for speakers of `standard' dialects but only 70\% accuracy for AAVE speakers, this performance gap -- if systematic and persistent -- moves beyond technical limitation to constitute disrespect by effectively signaling that certain voices are less worthy of accurate recognition. Disrespect occurs when people are reduced to mere category exemplars rather than being recognized as individuals with unique identities and equal moral worth. In the context of speech technologies, disrespect manifests when ASR systems consistently fail to recognize certain dialects or accents, effectively signaling that these speech patterns -- and by extension, the speakers themselves—are less worthy of technological accommodation.

We propose that traits used in algorithmic decision-making carry different moral weights depending on three key characteristics:

\begin{enumerate}
\item \textbf{Visible and unchangeable:} Traits that are immediately apparent and effectively immutable, such as race or certain congenital speech characteristics. These pose the highest risk for discriminate\textsubscript{2} because individuals cannot reasonably avoid discrimination based on these characteristics.
\item \textbf{Not immediately visible, possibly changeable with difficulty:} Traits like socioeconomic status, education level, or certain speech patterns that may serve as proxies for protected attributes (characteristics that correlate with race, gender, socioeconomic status, or other legally protected categories). While theoretically changeable, they often require substantial resources or effort to modify, and may perpetuate indirect discrimination \cite{barocas2016big, hellman2018indirect}.
\item \textbf{Not visible and more readily changeable:} Traits related to voluntary behaviors or temporary conditions that don't typically carry historical stigma. Examples might include temporary voice changes due to illness, deliberately adopted speaking styles, or context-specific vocal adjustments. Discrimination based on these traits, while potentially unfair, rarely produces the compounding injustice associated with categories (1) and (2).
\end{enumerate}

Speech patterns and accents often straddle the boundary between categories (1) and (2) -- they are immediately audible (like race is visible) and, for many speakers, are deeply intertwined with cultural identity and not readily changeable without significant personal cost. When ASR systems consistently misrecognize certain dialects, they effectively engage in discriminate\textsubscript{2} by reinforcing the marginalization of speech communities that have historically been deemed ``non-standard'' or less prestigious.

\begin{figure*}[t]
    \centering
    \includegraphics[width=\textwidth]{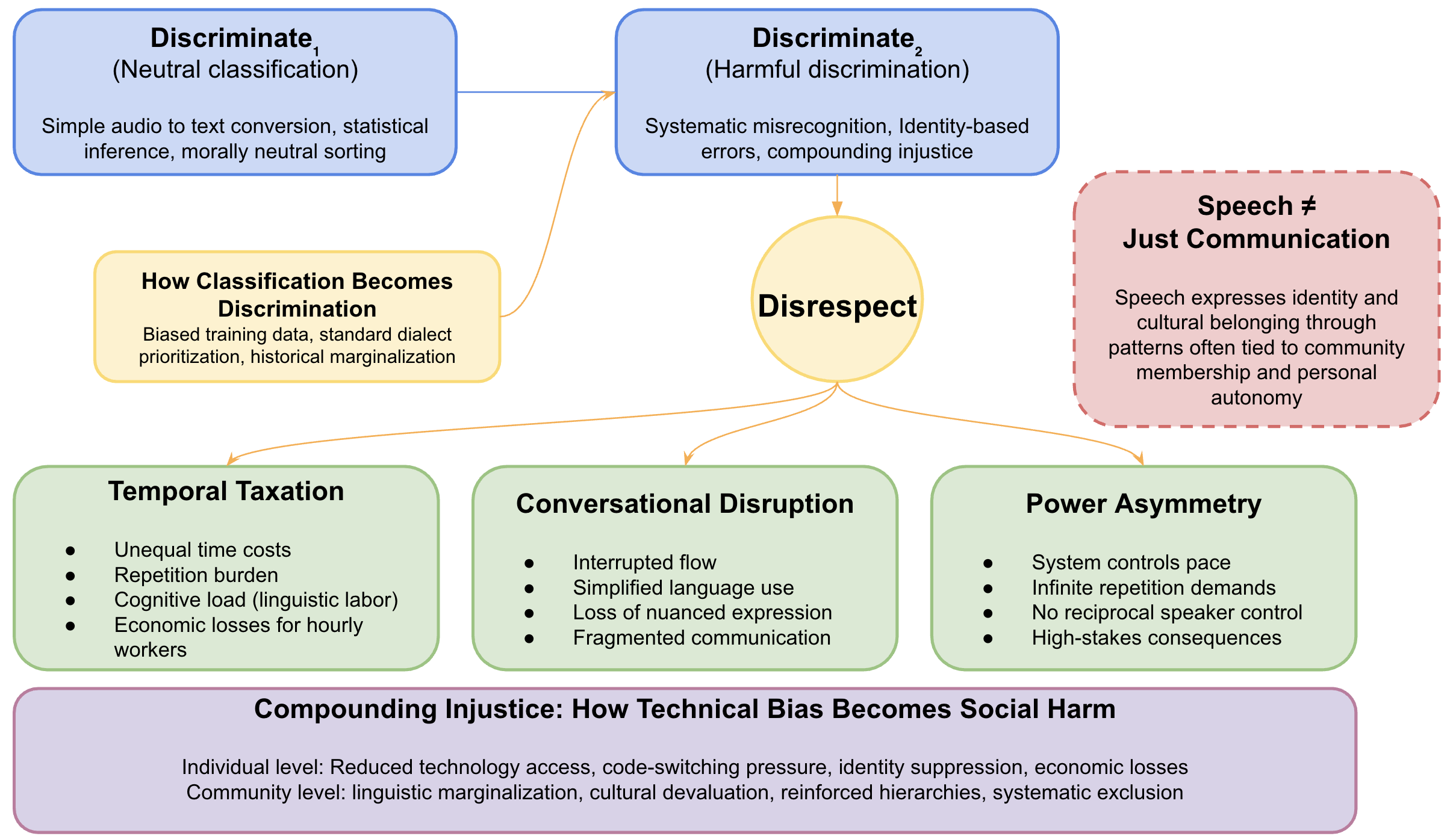}
    \caption{This diagram illustrates the central argument of our philosophical framework. Automatic speech recognition systems necessarily engage in discriminate\textsubscript{1} (morally neutral classification), but this process becomes discriminate\textsubscript{2} (harmful discrimination) through biased training data and prioritization of ``standard'' dialects. When systematic misrecognition concentrates among marginalized speech communities, it constitutes disrespect that manifests through three unique temporal dimensions: unequal time costs imposed on certain speakers (temporal taxation), disruption of natural conversational flow, and asymmetric power relationships between speakers and systems. These individual harms compound into broader social injustices at both personal and community levels. The side panel emphasizes that speech patterns are not merely communication tools but fundamental expressions of identity and cultural belonging, making recognition failures particularly harmful for speaker autonomy and dignity.}
    \label{fig:diagram}
\end{figure*}

This framework helps explain why seemingly neutral technical decisions in ASR -- such as prioritizing majority dialects in training data or optimizing for speakers who already enjoy high recognition rates -- can constitute ethical harms that extend beyond mere statistical disparities. By focusing on disrespect rather than just performance metrics, we can identify when ASR systems fail to treat all speakers as equally deserving of accurate recognition, regardless of their linguistic background. The relationships between these philosophical concepts and their manifestation in ASR bias are illustrated in Figure \ref{fig:diagram}, which shows how seemingly neutral technical classification transforms into harmful discrimination through specific mechanisms, ultimately producing three unique dimensions of temporal harm.

\section{The Ethical Challenges of Discrimination in ASR}

\subsection{ASR's Unique Ethical Terrain}

Having established this theoretical framework, we now examine how these dynamics play out specifically in the temporal domain, where ASR bias creates unique forms of harm that distinguish speech recognition from other algorithmic systems. In many discussions of discrimination -- credit scoring, university admissions, parole decisions -- one can track clear, binary outcomes (e.g., approval vs. rejection). Disparities in such direct decision points often trigger immediate scrutiny because they clearly embody \emph{discriminate\textsubscript{2}}. By contrast, ASR is typically operates upstream in decision-making pipeline simply converting audio to text. On the surface, this process exemplifies \emph{discriminate\textsubscript{1}} --  a morally neutral sorting mechanism that translates audio signals into language tokens. Yet these translations can still spawn \emph{discriminate\textsubscript{2}} if certain speech patterns are misrecognized or mistranscribed at systematically higher rates. 

Consider a hiring tool that uses ASR to parse video interviews. When the system produces error-filled transcripts for speakers with strong regional accents or non-native pronunciation, it inadvertently undermining their prospects. Although the system isn't overtly labeling them as inferior, the flawed transcript effectively serves as a hidden mark against these applicants. Similar dynamics appear in educational contexts, where non-native students using language tools to refine their English may be unfairly flagged for ``AI-generated'' content, even when their intent was simply to overcome linguistic barriers. 

These challenges intersect with emerging legal safeguards. New York City's Local Law 144 \cite{nyclaw144} mandates fairness audits of automated employment decision tools, while the Americans with Disabilities Act \cite{ada1990} offers protections for individuals with speech patterns tied to disabilities or medical conditions. If biased ASR systems systematically penalize such speakers, their deployment may violate civil rights laws \cite{eeoc2008}, transforming technical limitations into legally actionable discrimination.

\subsection{The Temporal Dimensions of Speech Recognition Bias}

Unlike text-based algorithmic systems, ASR operates in the temporal domain, creating unique ethical considerations that transcend simple performance metrics. The temporal nature of speech introduces several critical dimensions of potential harm that collectively distinguish speech recognition bias from other forms of algorithmic discrimination.

First, ASR errors create what we might call ``temporal taxation'' -- an unequal distribution of time costs that compounds existing social inequalities. When ASR misrecognizes speech, the burden of correction falls disproportionately on speakers whose accents or dialects diverge from the system's training data. To understand the magnitude of this burden, consider the quantitative implications: \citet{koenecke2020racial} found that ASR systems exhibit approximately twice the Word Error Rate (WER) for African American speakers compared to white speakers. If we conservatively estimate that each recognition error requires 10-15 seconds to identify, repeat, and verify correction, a five-minute customer service call could impose an additional 2-3 minutes of uncompensated labor on AAVE speakers simply to achieve the same outcome as speakers of "standard" dialects.

This temporal burden scales dramatically when we consider daily interactions. Consider a professional who makes 20 voice-based interactions daily -- such as a customer service representative using voice-recognition software -- who might lose 30-45 minutes per day to ASR-induced repetition and clarification. Annually, this could represent over 180 hours of additional labor, equivalent to more than four work weeks. For hourly workers using voice-based systems for job tasks, this temporal taxation would directly translate to economic losses through reduced productivity metrics or extended unpaid time completing required interactions.

The burden extends beyond mere time loss. Speakers from marginalized linguistic communities must interrupt their natural flow, code-switch to more ``standard'' pronunciations\footnote{This technological accommodation differs from natural code-switching, which serves communicative and social functions, in that it represents a unidirectional adaptation to overcome system limitations rather than mutual linguistic negotiation.}, simplify their vocabulary, and constantly monitor whether their adaptations successfully communicate their intent. This cognitive load represents what we might term ``linguistic labor'' -- the mental effort required to translate one's natural speech into forms legible to biased systems. Unlike speakers of privileged dialects who can engage with ASR systems automatically and unconsciously, marginalized speakers must maintain constant metalinguistic awareness, transforming routine interactions into sites of effortful performance.

Second, misrecognition disrupts the conversational flow that speech inherently requires to maintain coherence. The temporal rhythm of human conversation relies on precise timing -- pauses carry meaning, interruptions signal urgency, and the smooth exchange of turns enables complex ideas to unfold progressively. When ASR systems repeatedly interrupt to request clarification or produce errors that derail the conversation's trajectory, they don't merely misclassify individual words -- they fragment the speaker's ability to construct and convey complex thoughts.

This fragmentation likely produces measurable behavioral changes. Research on human-computer interaction suggests that when faced with repeated recognition failures, users adapt by simplifying their language \cite{porcheron2018voice}. For speakers experiencing high error rates, we can reasonably expect shortened utterances, reduced syntactic complexity, and abandonment of nuanced expression. A complaint about billing errors that might naturally unfold as a narrative explanation becomes reduced to stilted keywords: ``BILL. WRONG. FORTY. DOLLARS. JANUARY.'' The resulting interactions become stripped of the very features that make human communication rich and effective.

Third, the temporal aspects of ASR create fundamentally asymmetric power relationships between speakers and systems. The system controls the pace of interaction—it can interrupt at will, demand infinite repetitions, or simply fail to proceed until the speaker conforms to its expectations. The speaker, conversely, has no reciprocal power to slow the system down, request alternative recognition modes, or negotiate the terms of interaction. This asymmetry becomes particularly acute in high-stakes, time-sensitive contexts.

Consider customer service systems: when efficiency matters for business metrics, representatives relying on ASR to transcribe customer interactions may miss critical information from callers with non-standard accents, leading to extended call times and customer frustration. The caller cannot pause to negotiate better recognition; they must either persist through potentially fatal delays or hope human intervention overrides the system. In medical contexts, where voice-based documentation systems increasingly mediate patient -- provider interactions, ASR errors don't just waste time -- they risk introducing clinical errors when patients' symptom descriptions are mistranscribed. The temporal pressure in these contexts transforms recognition failures from inconveniences into potential catastrophes.

Employment interviews conducted through AI-mediated platforms present another stark example. When ASR systems consistently interrupt certain speakers for clarification, they don't just extend interview duration—they fundamentally alter the interaction's power dynamics. A candidate forced to repeat themselves appears less confident, their responses seem disjointed, and the natural rapport-building that occurs in smooth conversation becomes impossible. The system's temporal control thus translates directly into reduced employment opportunities for speakers of marginalized dialects.

These temporal dimensions -- taxation, disruption, and power asymmetry -- reveal how technical performance disparities translate into lived experiences of inequality. Standard fairness metrics that focus solely on accuracy rates fail to capture these temporal dynamics. A system might achieve statistical parity in eventual task completion while still imposing dramatically unequal temporal burdens. Two speakers might both successfully order a pizza through a voice assistant, but if one completes the task in 30 seconds while the other requires three minutes of repetition and clarification, the apparent equality masks a deeper form of discrimination.

Understanding these temporal dimensions suggests new directions for fairness metrics. Rather than measuring only final accuracy as a single numeric value, we should assess time-to-task-completion equity, conversational flow preservation, and linguistic labor requirements across different speaker populations. These metrics would reveal the hidden costs that current ASR systems impose on marginalized communities -- costs that accumulate into systematic disadvantages in a world increasingly mediated by voice interfaces.

\subsection{The Role of Inductive Bias in ASR}

ASR systems exemplify the inductive reasoning process described in Section \ref{sec2}: they learn acoustic-phonetic relationships from corpora of spoken language and apply these patterns to new inputs. These disparities stem not from conscious decisions to discriminate but from \emph{discriminate\textsubscript{1}} gone awry -- the model follows patterns most prominent in its training data, which typically under-represent minority dialects. Once encoded in the model, these patterns become self-reinforcing as the system treats ``standard'' speech as the default and other linguistic varieties as anomalies. This technical process aligns closely with Hellman's concept of compounding injustice: if a demographic group has historically faced sociolinguistic marginalization through stereotypes about ``unintelligible'' accents, then an ASR system that consistently misrecognizes that group's speech amplifies existing disadvantages, creating additional layers of social and economic harm.

\subsection{Disrespect in ASR Design and Deployment}

Speech patterns occupy a unique position on the spectrum of traits vulnerable to discrimination. While sometimes viewed as more malleable than race, in practice many people cannot simply ``speak differently''; accent, dialect, and linguistic style are deeply intertwined with cultural identity and community belonging \cite{lippi-green2012english, schiffrin1996narrative}. Speech carries conspicuous markers that can signal race, ethnicity, or social class \cite{armstrong2022sociolectal}, and these markers often coincide with longstanding social biases.

From the perspective developed in Section \ref{sec2.3}, speech often straddles the line between Category (1) traits (visible, unchangeable, historically oppressed) and Category (2) traits (less visible, theoretically changeable but with significant difficulty). When ASR systems consistently perform worse on certain dialects, especially those historically stigmatized as ``uneducated'' or ``broken English,'' they effectively treat these speech patterns as less legitimate, transforming \emph{discriminate\textsubscript{1}} into \emph{discriminate\textsubscript{2}}. Users face an unjust choice: code-switch, invest significant resources in accent modification, or remain marginalized by technologies that fail to accommodate their natural speech patterns.

The resulting harm extends beyond mere technical errors to symbolic disrespect—the implicit message that certain ways of speaking, and thus certain identities, are not legitimate enough to merit accurate transcription. This disrespect manifests when design choices prioritize data collection from ``standard'' dialects, when performance disparities lack transparency, and when the burden of adaptation falls on already marginalized speakers. Even without deliberate intent, ASR systems trained on biased data can reproduce and amplify existing social hierarchies \cite{lippert2023using}, effectively telling certain users: ``Our technology is not built with you in mind.''

Unlike some protected attributes, speech recognition could theoretically be improved through intentional technical interventions -- diverse training corpora, dialect-specific tuning, and transparent performance reporting across linguistic varieties. Yet realizing this promise requires more than improved aggregate accuracy metrics; it demands recognition of ASR's temporal dimensions and their unequal distribution of burdens, along with deliberate efforts to embed inclusivity throughout the technology's development cycle. Without such interventions, ASR risks not only perpetuating but temporally amplifying existing inequalities through the micropolitics of whose time and attention is valued in human-computer interactions.

\section{Speech: A Reflection of Identity and Autonomy}

\subsection{Voice as a Personal and Social Construct}

Human speech is remarkably diverse and multifaceted—far from the uniform phenomenon often assumed in technical discussions of ASR. Each person's speech pattern comprises a unique constellation of phonetic, prosodic, and linguistic features, forming what linguists term an ``idiolect'' \cite{coulthard2004author}. These distinctive vocal signatures emerge from the complex interplay of physiological, sociocultural, and temporal factors that shape human communication.

At the physiological level, voice quality, timbre, and articulation patterns vary based on physical characteristics and may change with aging or health conditions \cite{smith2018communication, davis1979acoustic}. Speech impairments -- whether congenital, acquired, or temporary -- produce distinctive acoustic patterns that can substantially differ from typical speech norms \cite{kent1979acoustic, pribuisiene2006perceptual, rojas2020does}. These variations represent not deviations from a norm but the natural diversity of human vocal production.

The sociocultural dimensions of speech run even deeper. Accents, dialects, and speech styles reflect membership in geographic, ethnic, or social communities, serving as deep markers of cultural identity and community belonging rather than merely superficial variations. Even within a single speaker, speech varies contextually through code-switching between dialects or adjusting prosodic features to match social situations \cite{genesee1982social}. This flexibility demonstrates speech's role as a dynamic social tool through which speakers navigate complex cultural landscapes.

Furthermore, speech patterns evolve throughout a person's life as they navigate different social environments, educational experiences, and professional contexts. This temporal development makes speech a continuously evolving aspect of identity rather than a static attribute. The dynamic nature of speech reflects the ongoing negotiation between individual expression and social participation.

When ASR systems fail to accommodate this rich variability, they effectively reduce speech to a narrow slice of ``standard'' features. Beyond generating transcription errors, these failures can erode users' sense of self-expression and authenticity, particularly when individuals feel pressured to modify their natural speech patterns to be understood by technology. The harm extends beyond inconvenience to touch on fundamental questions of identity and belonging in an increasingly voice-mediated world.

\subsection{Misrecognition vs. Misappropriation in Speech Technologies}

In examining the ethical implications of speech technologies, we must distinguish between two related but distinct phenomena that affect speakers in different ways. Misrecognition occurs when ASR systems systematically fail to accurately transcribe speech from certain dialects, accents, or speech patterns, effectively rendering these varieties technologically ``illegible.'' This problem -- our primary focus in this paper -- creates barriers to technology access and reinforces linguistic hierarchies by implicitly devaluing non-standard speech varieties. When systems cannot understand certain ways of speaking, they send a clear message about whose voices matter in the digital realm.

In contrast, misappropriation emerges in speech synthesis and voice generation technologies when systems produce speech that mimics or appropriates accents, dialects, or vocal characteristics without proper understanding of their cultural significance or without consent from the communities represented. While beyond our current scope, misappropriation raises complementary concerns about voice ownership, consent, and respectful representation that merit separate investigation. The rise of voice cloning technologies makes these questions increasingly urgent.

Both phenomena intersect with questions of identity and autonomy, but in different ways: misrecognition denies recognition of one's authentic voice, while misappropriation potentially exploits or misrepresents it. The fundamental ethical concern in both cases involves respect for linguistic diversity and individual expression. Understanding this distinction helps clarify the specific harms that arise when ASR systems fail to accommodate diverse speech patterns, as opposed to when synthetic voice technologies appropriate those patterns without permission.

\subsection{Autonomy and the Multi-Dimensional Nature of Speech}

Voice is integral not just to communication but also to one's autonomy -- the capacity to navigate social spaces on one's own terms. Speech carries paralinguistic dimensions that written text cannot capture, creating layers of meaning that extend far beyond the literal content of words. Intonation patterns, stress placement, rhythm, and speech rate convey emotional states, emphasis, irony, and other nuances essential to a speaker's intended meaning \cite{schuller2013computational}. These prosodic features transform identical word sequences into vastly different messages: a simple "yeah, right" can express agreement, sarcasm, or disbelief depending entirely on its vocal delivery.

Through phonetic variation, speakers also project identity and navigate social hierarchies. Subtle vocal choices -- whether conscious or unconscious -- signal group affiliation, express authority or deference, and perform various social identities. A speaker might adopt more formal pronunciation in professional settings while using relaxed articulation among friends, demonstrating how voice serves as a flexible tool for social positioning. Additionally, speakers naturally modulate timing, pauses, and turn-taking to control conversational flow -- capabilities that become disrupted when ASR systems misinterpret speech and force corrections.

When ASR systems consistently misinterpret these phonetic and prosodic cues -- or fail entirely to capture them -- they undermine a speaker's communicative autonomy. This challenge becomes particularly acute when certain accents or speech profiles are already socially stigmatized. Speakers may feel compelled to adopt more standardized pronunciation or suppress distinctive phonetic features to minimize recognition errors, engaging in a form of technological code-switching that exacts cognitive and emotional costs.

Such accommodations might seem trivial at first glance -- just ``speak more clearly'' -- but they represent a deeper compromise of authenticity and self-determination. The request to modify one's speech for technological legibility echoes historical demands that marginalized communities abandon their linguistic heritage to participate in mainstream society. Moreover, when misrecognitions disproportionately affect groups who already face sociolinguistic discrimination \cite{cocchiara2016sounding, kristiansen2001social}, they compound existing injustices, further restricting already-limited social and professional opportunities. In this way, ASR systems risk becoming another mechanism through which linguistic prejudice operates, now encoded in algorithms rather than human bias alone.

\subsection{Linguistic Pluralism vs. Standardization}

A fundamental tension in ASR development lies between linguistic pluralism and standardization. Most commercial systems optimize for a presumed ``standard'' -- typically the dialect associated with educated, middle/upper-class speakers from dominant social groups. This approach produces higher accuracy for majority speakers but systematically disadvantages linguistic minorities. This technical choice reflects broader language ideologies -- systems of beliefs about which language varieties are ``correct,'' ``professional,'' or ``intelligible'' \cite{woolard2020language}. Under standard language ideology, one dominant dialect is presumed superior or more appropriate \cite{milroy2001language}, despite linguistic evidence that all dialects are equally valid as systematic communication systems \cite{labov1972language, rickford2007spoken}.

The recent executive actions declaring English as the official language of the United States \cite{whitehouse2025english} exemplify how standard language ideology operates at the highest levels of policy, providing stark illustration of the political dimensions underlying ASR bias.\footnote{On March 1, 2025, Executive Order 14224 declared English the official language of the United States federal government. While primarily affecting government communications, its ideological impact extends to technology development as companies may interpret official monolingualism as license to deprioritize linguistic diversity.} This governmental endorsement of linguistic hierarchy amplifies the harms we identify in ASR systems. When the state officially designates certain languages as legitimate and others as foreign, it provides ideological cover for technological systems that similarly privilege ``standard'' English while marginalizing linguistic diversity. Such policies create what we might call a ``cascade of legitimation'' for linguistic discrimination -- if English is the only `official' language, then ASR systems that fail to recognize non-standard English dialects -- not just other languages -- can frame their limitations as alignment with national policy, even though these varieties remain forms of English.

When standard language ideology influences ASR design, developers prioritize the acoustic patterns of ``standard'' varieties \cite{koenecke2020racial}, with less attention to systematic inclusion of diverse speech patterns. These design decisions constitute a form of language policy \cite{spolsky2009language, markl2022language}, tacitly endorsing existing sociolinguistic hierarchies. The consequences extend beyond mere convenience -- they affect which voices are deemed worthy of technological accommodation and, by extension, which speakers can fully participate in increasingly voice-mediated digital environments. The convergence of official language policies with biased ASR systems threatens to create technologically mediated mechanisms for linguistic exclusion that operate with both state sanction and algorithmic efficiency.

A pluralist approach would explicitly value diverse speech varieties throughout the development process. This approach recognizes non-standard varieties not as deficient but as legitimate linguistic systems carrying cultural significance and personal identity. The technical challenges are substantial -- gathering representative speech samples from marginalized communities, designing models that handle code-switching and dialectal variation, and developing evaluation metrics that account for linguistic diversity. Yet these challenges are fundamentally ethical as well as technical, returning us to the concept of disrespect introduced in Section \ref{sec2.3}. When ASR systems systematically fail to recognize certain speech varieties, they effectively treat these speakers as less worthy of technological accommodation.

The distinction between standardization and pluralism manifests throughout the ASR development pipeline in concrete design choices. During data collection, developers must decide which speech varieties to sample, in what proportions, and with what level of phonetic diversity. These decisions shape the fundamental capabilities of the resulting systems. In acoustic modeling, the treatment of phonetic variants reveals underlying values -- are variations treated as errors to be corrected or as legitimate alternatives to be recognized? The evaluation process similarly embeds values about whose speech matters: systems tested primarily with speakers of ``standard'' varieties will inevitably perform better for those populations while failing others. Finally, error handling strategies reveal assumptions about adaptation -- should systems learn to understand diverse speakers, or must speakers modify their speech to be understood?

Each of these decisions encodes values about which speech forms matter, with profound implications for speaker autonomy and linguistic equality. In an era where linguistic nationalism gains official status, these design choices take on new political significance. ASR systems can either resist linguistic hegemony by ensuring technological access across language varieties, or they can become instruments of enforcement, automating the exclusion that official language policies initiate. As speech interfaces become increasingly central to digital participation, the question of whose voices are recognized becomes inseparable from whose identities are respected and whose autonomy is preserved.

\section{Implications for Law and Policy}

Our philosophical framework reveals critical gaps in current legal approaches to ASR bias. Existing anti-discrimination law typically requires proving disparate impact through statistical disparities, yet our analysis shows that temporal taxation and identity-based harms may occur even when aggregate accuracy appears equitable. 

Current fairness audits focus primarily on outcome disparities such as differential WERs across demographic groups. However, our framework suggests regulators should expand their scope to measure temporal burdens -- quantifying how much additional time and effort different groups must expend to achieve the same outcomes. This expanded understanding of algorithmic harm could inform new standards under emerging regulatory frameworks like New York City's Local Law 144 \cite{nyclaw144}, pushing beyond simple accuracy metrics to capture the lived experience of interacting with biased systems.

The question of accommodation presents another critical challenge. The Americans with Disabilities Act \cite{ada1990} requires ``reasonable accommodation'' for disabilities but typically places the burden on individuals to request specific modifications. Our analysis suggests that for speech technologies, this burden-allocation may be fundamentally misplaced. Just as buildings must be wheelchair accessible by default rather than modified upon request, ASR systems should proactively accommodate linguistic diversity as a baseline requirement. This shift from reactive to proactive accommodation would recognize that linguistic variety is not a deviation requiring special treatment but a fundamental aspect of human communication deserving universal design.

Perhaps most provocatively, our compounding injustice framework implies that using speech data from marginalized communities without ensuring equitable system performance constitutes a form of extraction. When companies harvest linguistic data from AAVE speakers to improve overall system robustness (often benefiting primarily 'standard' speakers) but fail to ensure those same AAVE speakers can effectively use the resulting systems, they engage in a form of algorithmic colonialism -- extracting value while denying reciprocal benefit. This dynamic suggests the need for new data governance models that recognize linguistic data sovereignty, giving speech communities meaningful control over how their vocal patterns are collected, used, and commodified. Such frameworks would move beyond simple consent models to ensure that data contribution translates into equitable system access.

These legal and policy implications demonstrate why philosophical analysis matters: without understanding ASR bias as a form of disrespect that compounds historical injustices, regulatory responses risk addressing symptoms while leaving fundamental power asymmetries intact.

\section{Conclusion}

Our analysis positions ASR within a broader moral and political framework, arguing that entrenched biases in speech technology are neither incidental nor remediable by technical optimizations alone. By introducing the notion of disrespect and linking it to the harms inflicted upon speakers of marginalized dialects, we illuminate how ASR can perpetuate entrenched inequalities and social stigmas. In doing so, we shift focus from mere error rates to the ethical implications of whose voices are systematically misrecognized.

The unique temporal dimensions of speech recognition bias -- temporal taxation, disruption of conversational flow, and asymmetric power dynamics -- further distinguish ASR from other algorithmic systems. These temporal burdens fall disproportionately on speakers whose linguistic identities deviate from the presumed ``standard,'' compounding existing social inequalities through the micropolitics of time and attention in human-computer interaction.

The critique of inductive reasoning in ASR -- where historical data imbalances become self-reinforcing -- echoes concerns long articulated in philosophy about how generalizations can slide from practical heuristics (\emph{discriminate\textsubscript{1}}) to morally charged prejudices (\emph{discriminate\textsubscript{2}}). Under conditions of compounding injustice, consistent errors in transcribing certain voices exceed the threshold of mere technological shortcomings. They instead signal deeper forms of moral disregard, suggesting that the traits and linguistic identities of underrepresented communities are less worthy of accurate technological recognition.

Responding ethically to ASR bias, therefore, involves transcending narrow fixes such as targeted data-collection or model updates, although these remain vital. It requires an explicit commitment to linguistic pluralism, wherein diverse dialects and accents are treated not as problematic deviations but as legitimate speech forms necessitating fair representation. This commitment should manifest in concrete research directions: developing evaluation metrics that capture not just aggregate accuracy but the distribution of temporal burdens across speaker groups; creating transparent benchmarking standards that assess performance across diverse speech communities; designing interfaces that mitigate power asymmetries by giving speakers greater control over correction and adaptation processes; and establishing data governance frameworks that ensure traditionally marginalized speech communities maintain authority over how their linguistic data is collected and used.

Moreover, clarifying the philosophical underpinnings of discrimination in speech technology can inform more accountable AI governance. Far from being a peripheral concern, ensuring that ASR systems acknowledge the autonomy and dignity of all speakers is essential for fostering genuinely fair, accountable, and transparent uses of machine learning. By weaving together ethical analysis, technical design, and policy considerations, the field can advance toward ASR systems that affirm rather than undermine the linguistic identities of the communities they purport to serve.

As voice interfaces increasingly mediate access to essential services and opportunities, the question of whose speech patterns are recognized becomes inseparable from questions of equality and justice. Speech is not merely a communication medium but a fundamental expression of identity and community belonging. When ASR systems systematically fail to accommodate linguistic diversity, they don't just produce technical errors -- they participate in a form of structural disrespect that echoes and reinforces broader patterns of social marginalization. Addressing this challenge requires not only technical innovation but a deeper commitment to recognizing the moral worth and linguistic autonomy of all speakers.

\section*{Acknowledgments}
We thank Daniel Susser and Corey Miller for their valuable comments.

\bibliography{aaai25}

\end{document}